\documentclass[letterpaper, 10 pt, conference]{ieeeconf}  

\IEEEoverridecommandlockouts                              

\let\labelindent\relax

\usepackage{etoolbox} 
\usepackage[utf8]{inputenc} 
\usepackage[T1]{fontenc}    
\usepackage{fonttable}
\usepackage[english]{babel} 
\usepackage[protrusion=true,expansion=true]{microtype}
\usepackage{comment}
\usepackage{cancel}
\usepackage{csquotes}
\usepackage{listings}
\usepackage{nameref}
\usepackage{paralist}
\usepackage{xspace}
\usepackage{setspace}
\usepackage{makeidx}
\usepackage{pdfpages}
\usepackage{times}
\usepackage{microtype}
\usepackage{soul}

\usepackage{amsmath,amssymb} 
\usepackage{amsfonts}       
\usepackage{mathrsfs} 
\usepackage{mathtools}
\usepackage{array} 
\usepackage{nicefrac}       
\usepackage{breqn}          
\usepackage{textcomp}
\usepackage{bm} 
\usepackage{pifont} 
\usepackage[
  separate-uncertainty = true,
  multi-part-units = repeat
]{siunitx}

\usepackage{graphicx}
\usepackage{wrapfig}
\usepackage{adjustbox} 
\usepackage{epsfig}

\usepackage{float}
\usepackage{stfloats}
\usepackage{placeins}
\usepackage{subcaption}
\usepackage[font=small,labelfont=bf]{caption}
\usepackage{lscape}      

\usepackage{tikz}  
\usepackage{makecell}
\usepackage{overpic}
\usepackage{cuted} 

\newsavebox{\subfigbox}

\usepackage{algorithm}
\usepackage[noend]{algpseudocode}

\usepackage{array} 
\usepackage{tabularx}
\usepackage{multirow}
\usepackage{multicol}
\usepackage{booktabs}   
\usepackage{tablefootnote}
\usepackage{threeparttable}

\usepackage{paralist}
\usepackage{enumitem}
\setlist[itemize]{noitemsep,leftmargin=*,topsep=0in}
\setlist[enumerate]{noitemsep,leftmargin=*,topsep=0in}

\usepackage{url}            
\PassOptionsToPackage{table}{xcolor}
\usepackage{color,colortbl} 
\usepackage{soul}

\PassOptionsToPackage{capitalize, noabbrev}{cleveref}

\usepackage{hyperref}

\usepackage{blindtext}
\usepackage{bbm}

\usepackage{lipsum}

\usepackage{titlesec}
\titlespacing{\section}{0pt}{0.3\baselineskip}{0.25\baselineskip}
\titlespacing{\subsection}{0pt}{0.2\baselineskip}{0.15\baselineskip}
\titlespacing{\subsubsection}{0pt}{0.05\baselineskip}{0.03\baselineskip}

\renewcommand{\paragraph}[1]{\noindent\textbf{#1} --}

\newcommand{\cross}{\textcolor{red}{\ding{55}}}
\newcommand{\tick}{\textcolor{teal}{\ding{51}}}

\definecolor{color1}{rgb}{0.6, 0.4, 0.05}
\definecolor{color2}{rgb}{0.0, 0.7, 0.7}
\definecolor{color3}{rgb}{0.35, 0.75, 0.0}
\definecolor{color4}{rgb}{0.4, 0.8, 0.0}
\definecolor{color5}{rgb}{0.5, 0.0, 0.5}
\definecolor{color6}{rgb}{0.99, 0.84, 0.69}
\definecolor{color7}{rgb}{0.8, 0.3, 0.3}
\definecolor{color8}{rgb}{0.2, 0.4, 0.8}
\definecolor{color9}{rgb}{0.6, 0.2, 0.6}
\definecolor{revision_color}{rgb}{1,0,0}

\newcommand{\mycomment}[1]{#1} 

\usepackage{enumitem}

\newcommand{\modelName}{\textsc{Cobalt}\xspace}

\title{\modelName: Crowdsourcing Robot Learning via \\ Cloud-Based Teleoperation with Smartphones}

\author{
    Ayush Agarwal$^{1^*}$,
    Ansh Gandhi$^{1,2^*}$,
    Jeremy A. Collins$^{1}$,
    Omar Rayyan$^{3}$,
    Aryan Sarswat$^{1}$,
    Ranjani Koushik$^{1}$, \\
    Masoud Moghani$^{4}$,
    Ajay Mandlekar$^{5}$,
    Animesh Garg$^{1}$%
    \thanks{$^{*}$denotes equal contribution}%
    \thanks{$^{1}$Georgia Institute of Technology \quad $^{2}$University of California, Berkeley}%
    \thanks{$^{3}$New York University Abu Dhabi (NYUAD) \quad $^{4}$University of Toronto}%
    \thanks{$^{5}$NVIDIA}%
}

\begin{document}

\makeatletter
    \let\@oldmaketitle\@maketitle
    \renewcommand{\@maketitle}{\@oldmaketitle
    \centering
    \includegraphics[width=0.98\linewidth]{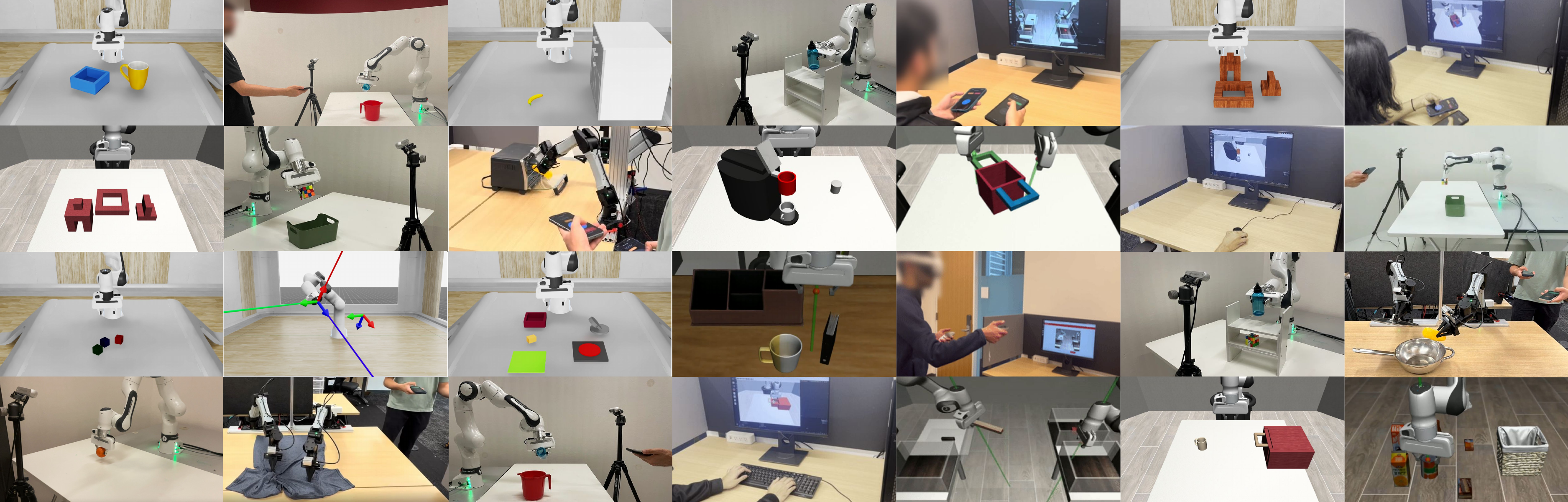}
    \captionof{figure}{\modelName can be used to collect data across a variety of both simulated and real-world environments, including bimanual tasks.\vspace{-0.5cm}}
    \label{fig:headliner}
    }
\makeatother

\maketitle
\thispagestyle{empty}
\pagestyle{empty}
\setcounter{figure}{1}

\newcommand{\ansh}[1]{\textcolor{color1}{\mycomment{ansh: #1}}}
\newcommand{\jeremy}[1]{\textcolor{color2}{\mycomment{jeremy: #1}}}
\newcommand{\aryan}[1]{\textcolor{color3}{\mycomment{aryan: #1}}}
\newcommand{\ayush}[1]{\textcolor{color4}{\mycomment{ayush: #1}}}
\newcommand{\kaavya}[1]{\textcolor{color5}{\mycomment{kaavya: #1}}}
\newcommand{\omar}[1]{\textcolor{color6}{\mycomment{omar: #1}}}
\newcommand{\ranjani}[1]{\textcolor{color7}{\mycomment{ranjani: #1}}}
\newcommand{\ag}[1]{\textcolor{color8}{\mycomment{animesh: #1}}}
\newcommand{\masoud}[1]{\textcolor{color9}{\mycomment{masoud: #1}}}

%






\vspace{-11mm}

\begin{abstract}
    The scarcity of large-scale, high-quality demonstration data remains a bottleneck in scaling imitation learning for robotic manipulation. We present \modelName, a teleoperation platform designed to democratize robot learning at scale both in simulation and in the real world. By leveraging vectorized environments, our scalable, load-balanced infrastructure supports concurrent teleoperation by multiple users on a single GPU, yielding a significant reduction in teleoperation cost. Operators can connect from nearly anywhere on Earth using commonly available devices, including single or dual smartphones, VR headsets, 3D mice, and keyboards. An in-memory data cache and efficient video streaming keep control and rendering synchronous, sustaining dozens of concurrent users at 20 Hz with sub-100 ms end-to-end latency for up to 8 concurrent users per GPU. We also demonstrate stable operation supporting 256 simulated clients across 8 GPUs, underscoring the system's ability to scale across hardware and within individual servers. We perform a comprehensive user study showing that phone-based teleoperation performs comparably to or better than specialized hardware, enabling faster, more ergonomic data collection. To ensure data quality, \modelName logs a suite of real-time metrics to automatically filter suboptimal demonstrations. We further demonstrate that a structured user training curriculum significantly improves data collection quality. Guided by insights from our user study, we crowdsource the collection of a large-scale, high-quality pilot dataset with 7500+ demonstrations (50+ hours) collected with smartphones across nine countries over five days. We validate the dataset's quality by training state-of-the-art imitation learning algorithms. Please visit \href{https://cobalt-teleop.github.io}{cobalt-teleop.github.io} for more details.
\end{abstract}

\section{Introduction}
\label{sec:introduction}

The long-term vision of robotics increasingly relies on data-driven methods like imitation learning \cite{chi2023diffusion, toolmanipulation, goyal2023rvt, lee2024behavior, mete2024quest, aloha}. While these techniques efficiently teach robots skills using human demonstrations, their ability to generalize remains severely limited by the quantity, quality, and diversity of available training data. This constitutes a fundamental bottleneck: compared to the billions of images and trillions of text tokens fueling foundation models in computer vision and NLP \cite{commoncrawl, laion}, robotics operates in a data desert, with even the largest datasets being orders of magnitude smaller ($\mathcal{O}(10^6)$ trajectories) \cite{brohan2022rt, droid, vuong2023open, walke2023bridgedata}. Bridging this vast data gap is arguably the most critical step toward realizing robots with broad capabilities for assisting humans across diverse tasks and environments.

Gathering demonstrations on physical hardware is notoriously time-consuming and cost-prohibitive, limiting dataset size and diversity. Collecting demonstrations in simulation serves as a complementary method of scaling robotics data. Modern simulation frameworks, such as MuJoCo \cite{mujoco} and Isaac Sim \cite{isaacsim}, accelerate dataset generation by enabling fast creation of diverse teleoperation environments \cite{telemoma, roboturk, mimicgen, robocasa}. Regardless of the simulated or physical teleoperation environment, a critical obstacle in bridging the data gap is how effectively human operators can be brought into the loop to provide high-quality demonstrations. This is heavily influenced by factors such as cost, ease of onboarding, ergonomics, and ease of use.


We present \modelName, a scalable data collection platform that leverages cloud-based infrastructure to enable teleoperation from geographically distributed users using a variety of off-the-shelf devices. \modelName is, to our knowledge, the first to support concurrent, uninterrupted teleoperation across GPU-accelerated vectorized simulation environments (supporting Isaac Lab \cite{orbit}, robosuite \cite{robosuite}, and LIBERO \cite{liu2023libero}), significantly improving the cost and efficiency of data collection. \modelName is globally deployable and supports remote operators connecting via single smartphones (Android/iOS), dual smartphones for bimanual control, VR headsets, 3D mice, and keyboards. We achieve low-latency (sub-$100$ ms at $20$ Hz) interaction through optimized networking, caching, and multiprocessing.

Recognizing that scale without quality is insufficient, \modelName incorporates a suite of real-time performance metrics to automatically filter suboptimal demonstrations and a structured training curriculum proven to improve user proficiency and data quality. Our contributions are summarized below:

\begin{enumerate}
    \item \textbf{\modelName}: An open-source, cloud-based teleoperation platform designed for scalability and accessibility, supporting 190+ environments. \modelName accommodates several concurrent users on a \textit{single GPU} and integrates several commonly available input devices, including smartphones, thus lowering the barrier to entry.
    \item \textbf{Data Quality at Scale}:
    We introduce a structured training curriculum and an extensive suite of performance metrics that help refine data quality and lay the groundwork for future systems capable of autonomous user onboarding and data curation at scale.
    \item \textbf{User Study and Analysis}: We perform a comprehensive user study comparing input devices for teleoperation, providing insights into device ergonomics and performance, alongside stress tests quantifying the platform's scalability.
    \item \textbf{Pilot Dataset}: We crowdsource a pilot dataset ($7500$+ human-collected demos, $50$+ hours) using \modelName from $50$+ \textit{inexperienced} teleoperators across \textit{nine} countries over \textit{five} days and evaluate its quality by training imitation learning policies.
\end{enumerate}

\section{Related Work}
\label{sec:related-work}

\begin{table*}[!t]
    \centering
        \caption{\textbf{Comparison of Existing Teleoperation Techniques in Literature}. \textbf{S} = Smartphone, \textbf{VR} = Virtual Reality, \textbf{3DM} = 3D mouse, \textbf{K} = keyboard, \textbf{J} = Joystick. $^*$TeleMoMa data collection infrastructure is not public. $^{**}$Assumes RTX 3090, real value varies per GPU. Assumes zero marginal cost of a smartphone.}
    \label{tab:approaches}
    \resizebox{\textwidth}{!}{
    \begin{tabular}{ccccccccc}
        \toprule
        \rowcolor[HTML]{FFEED4}
        Method & RoboTurk \cite{roboturk} & MoMaRT \cite{momart} & TeleMoMa \cite{telemoma} & RoboTurk Real-World \cite{roboturkirl} & GELLO \cite{gello} & ALOHA \cite{aloha} & \textbf{\modelName} \\
        \midrule
        Device Cost & \$0-\$500 & \$0-\$500 & \$0-\$500 & \$0-\$500 & $\sim$\$300 & $\sim$\$20k & \$0-\$500 \\
        \rowcolor[HTML]{EFEFEF}
        Input Devices & S, VR, 3DM, K & S, J & S, VR, 3DM, K & S & J & J & S, VR, 3DM, K \\
        Coverage & iOS & iOS & iOS & iOS & $-$ & $-$ & iOS, Android \\
        \rowcolor[HTML]{EFEFEF}
        Bimanual & \cross & \cross & \tick & \cross & \tick & \tick & \tick \\
        Simulators & MuJoCo & MuJoCo & MuJoCo & MuJoCo & $-$ & $-$ & MuJoCo \& Isaac Lab \\
        Training Curriculum & \cross & \cross & \cross & \cross & \cross & \cross & \tick \\
        \rowcolor[HTML]{EFEFEF}
        Sim/Real & Sim & Sim & Sim \& Real & Real & Real & Real & Sim \& Real \\
        Remote & \tick & \tick & \tick & \tick & \cross & \cross & \tick \\
        \rowcolor[HTML]{EFEFEF}
        Publicly Available & \cross & \cross & \tick $^*$ & \cross & \tick & \tick & \tick \\
        Cloud-Scaling & \tick & \cross & \cross & \cross & \cross & \cross & \tick \\
        \rowcolor[HTML]{EFEFEF}
        Users Per Machine & 1 & 1 & 1 & 1 & $-$ & $-$ & 8+$^{**}$ \\
        \bottomrule
    \end{tabular}
    }
\end{table*}

\subsection{Teleoperation Frameworks}

Large-scale data collection requires robust infrastructure that supports low-latency streaming, offers accessible input modalities, and enables distributed deployment.

RoboTurk \cite{roboturk} introduced a server-client architecture that shifted simulation computation to remote servers. This design allowed participants to control robots through smartphone and VR interfaces with minimal local hardware requirements, improving scalability and enabling crowdsourcing.

Subsequent works, such as TeleMoMa \cite{telemoma} and MoMaRT \cite{momart}, extended these contributions to mobile manipulators, enabling more complex tasks and control strategies. However, these platforms generally lack comprehensive user testing, quantitative evaluation of collected trajectories, and a demonstration of the ability to scale crowdsourced data collection.

\modelName differentiates itself from these works in its ability to arbitrarily scale geographically distributed data collection with increasing compute. Earlier efforts collected datasets from a handful of users and tasks \cite{telemoma, roboturk}, enabling users to gain strong proficiency and provide near-expert demonstrations over time. We demonstrate true global crowdsourced teleoperation, yielding over $50$ hours of successful demonstrations across $10$+ environments, multiple simulators, and $50$+ inexperienced teleoperators. In contrast to prior platforms \cite{telemoma, roboturk, momart}, \modelName supports concurrent, uninterrupted teleoperation across vectorized environments on a single GPU, significantly improving scalability, efficiency, and cost-effectiveness. To lower the barrier for community adoption and extension, \modelName will be released as a fully open-source teleoperation platform supporting accessible, low-cost devices like smartphones, including both Android and iOS. With around $80\%$ of the global market share \cite{nicolle2024android}, providing Android support is particularly crucial in enabling global crowdsourcing. \modelName overcomes significant limitations of previous systems that were either closed-source \cite{roboturk, momart, roboturkirl}, incompletely released \cite{telemoma}, or required costly specialized hardware \cite{aloha, gello}.

\vspace{-1mm}
\subsection{Input Devices}
\vspace{-1mm}

Scalable teleoperation depends on selecting input devices that strike a balance between cost, availability, user comfort, and precision. Specialized options, such as leader-follower setups \cite{aloha} or VR headsets \cite{arcap, dexhub}, can provide highly accurate and intuitive control, but often require equipment that is neither widely accessible nor fatigue-free over extended sessions. Similarly, 3D mice have been refined for teleoperation using techniques like deadbands and low-pass filtering \cite{3dmice}, but remain limited in their adoption due to cost and niche usage. By contrast, most modern smartphones include built-in augmented reality (AR) frameworks capable of tracking 6-DoF poses with competitive accuracy \cite{vslam}. This has enabled platforms such as RoboTurk \cite{roboturk} and MoMaRT \cite{momart} to successfully gather large volumes of manipulation data. Furthermore, independent work on user interface improvements, such as explicit input assistance \cite{fastexplicitteleop} and automated grasp planning \cite{graspgf}, further simplify the data collection process by reducing fatigue and user error. Overall, the broad availability of smartphones presents an ideal pathway to crowdsource demonstrations at scale, especially for tasks requiring full pose control, without specialized hardware.

\section{\modelName: Design and Architecture}
\label{sec:methods}

\modelName is a scalable, cloud-based data collection platform that enables users worldwide to remotely teleoperate simulated and real robots. By leveraging low-latency networking, diverse input devices, multiple simulation frameworks, and real-world teleoperation capabilities, \modelName enables large-scale crowdsourcing and democratizes the creation of high-quality robotics datasets.

\modelName improves scalability and accessibility, as compared to prior work in RoboTurk \cite{roboturk}, through the integration of robust cloud infrastructure and support for a variety of input devices. The platform accommodates smartphones (both Android and iOS), virtual reality (VR) headsets, keyboards, and 3D mice. For bimanual tasks, \modelName also supports VR headsets and dual smartphones. By offering an extensible control interface, \modelName allows developers to integrate arbitrary input devices. Additionally, the platform incorporates a structured training curriculum to onboard users effectively, ensuring the collection of high-quality demonstration data.

\begin{figure*}
  \centering
  \includegraphics[width=\linewidth]{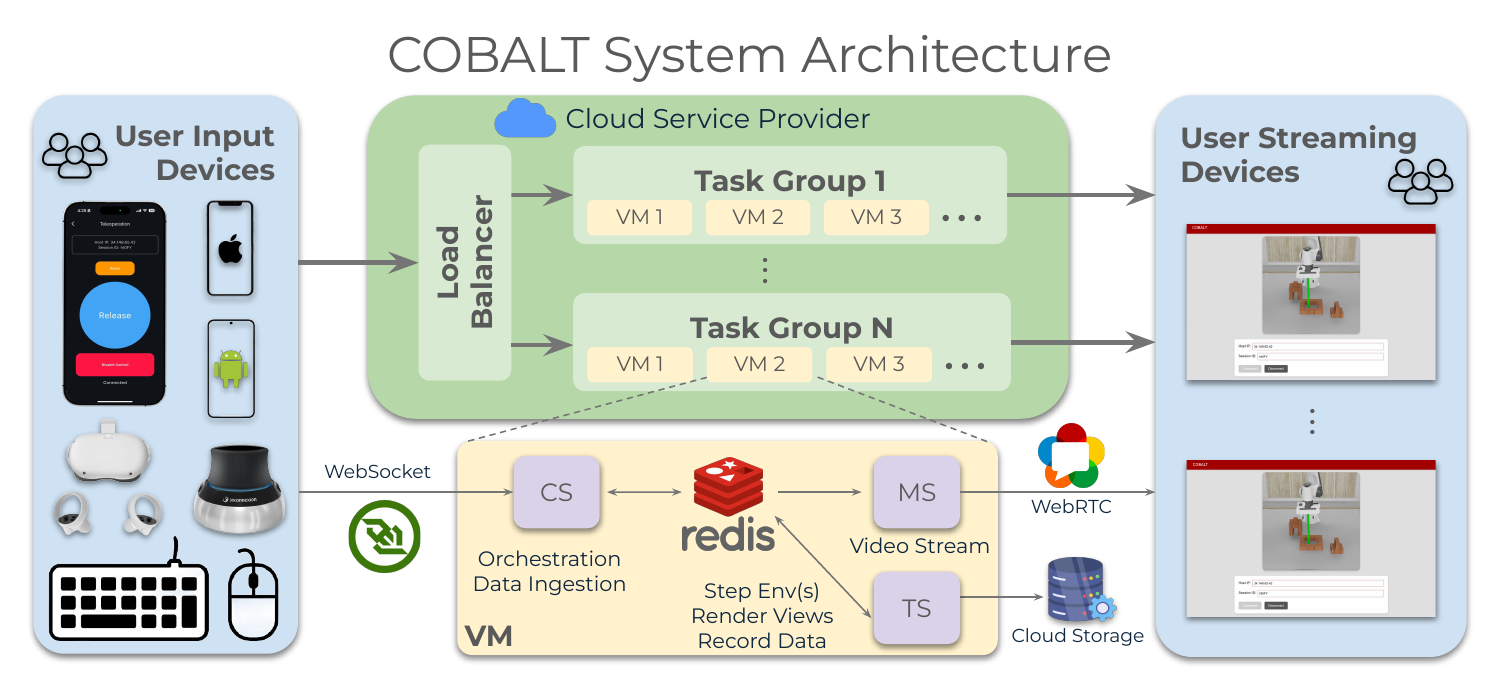}
  \caption{\modelName System Architecture. a) Cloud provider hosts one group of virtual machines (VM) per task, with dynamic allocation of servers based on demand. b) A load balancer sits in front of the different groups of servers, functioning as a rate limiter and reverse proxy. c) Three main services are utilized: CS (Client Session Service) for client data ingestion, MS (Media Service) for video streaming, and TS (Teleoperation Service) for the vectorized simulation backend. d) Data pipeline handles data storage, cleaning scripts, and augmentation.}
  \label{fig:method}
\end{figure*}

\subsection{\textbf{System Architecture}}

\modelName's architecture is designed to be intuitive and modular. Users connect to a server using their input device, and the server streams the simulated environment to a web browser or VR headset. Building such a system to scale, however, requires nuanced design choices. Namely: (1) users should be able to connect to and exit the system at any point in time, essentially as a \textit{distributed on-demand} service, (2) the platform must support concurrent users on a single simulation instance to optimize compute-cost trade-offs with maximal resource utilization, and (3) the system must support teleoperation with low latency so that users have a smooth and comfortable experience.

\modelName meets these requirements with a modular architecture to facilitate ease of development, use, and extensibility. \modelName consists of three primary components: a Client Session (CS) Service to communicate with client input devices, a Teleoperation Service (TS) to run a simulation backend capable of handling multiple users, and a Media Service (MS) to stream visual feedback with low latency (Figure~\ref{fig:method}).

\paragraph{Centralized Communication via Redis}
To handle communication among all three services within the platform, we utilize Redis, an in-memory database with low-latency read and write operations that can be deployed in a distributed fashion across several machines. This database decouples the main services, allowing them to exchange state information such as user commands and rendered video frames asynchronously and efficiently.

\paragraph{User Connection and Input Handling}
Users connect to the platform via the Client Session Service using a WebSocket connection from their chosen input device. This service manages user authentication and session lifecycles and acts as the primary ingestion point for user control data. It receives raw pose information from the client device at $20$ Hz and continuously publishes these 6-DoF pose commands to the Redis store, tagged by user session. This approach allows users to join and leave seamlessly on-demand. Redis enables the Client Session Service to asynchronously communicate this information to downstream services, maintaining performance and consistency.

\paragraph{Vectorized Simulation Core}
The Teleoperation Service is our system's computational engine. \modelName leverages \textit{vectorized simulation environments}, allowing the service to manage and step multiple independent simulation instances \textit{concurrently} on a \textit{single} GPU. The service orchestrates the assignment of available simulation environments to clients upon new connections. For each active user session, the service subscribes to the session's pose commands in the Redis store, pulls the latest pose update, performs necessary coordinate transformations, and dispatches the action to the corresponding simulation environment. Internally, the Teleoperation Service applies the standardized 6-DoF commands to the specific robot model within the chosen simulator backend. After stepping the simulation, it renders the visual output for each environment, encodes it with H.264, and publishes the encoded frames to a fixed-size buffer stored in Redis. In parallel, this service also logs all pertinent demonstration data (states, actions, timestamps, metrics) for offline use.

\paragraph{Low-Latency Visual Feedback}
To provide users with real-time visual feedback, the Media Service subscribes to the encoded video frames published to Redis by the Teleoperation Service. It utilizes WebRTC to establish a direct, low-latency peer-to-peer streaming connection with the user's display client (e.g., a web browser or VR headset). This minimizes the delay between a user's action and the visual result, which is crucial for real-time teleoperation.

\paragraph{Scalable Cloud Deployment}
The entire architecture is designed for robust deployment on cloud platforms. We containerize and deploy the system across auto-scaling VM instance groups, segregated by task type and/or geographic region. A central load balancer distributes incoming user connections, thus enabling high availability and responsiveness. This infrastructure allows our platform to dynamically scale compute resources based on demand, supporting in principle an arbitrary number of concurrent users globally while keeping operational costs low.

\subsection{\textbf{Training Curriculum}}
\label{subsec:curriculum}

To ensure users are prepared for teleoperating simulated robots using \modelName, we developed a training curriculum consisting of calibration and evaluation tasks. This curriculum is designed to onboard users to ensure they can collect high-quality demonstration data.

\begin{figure}[h]
    \centering
    \includegraphics[width=\linewidth]{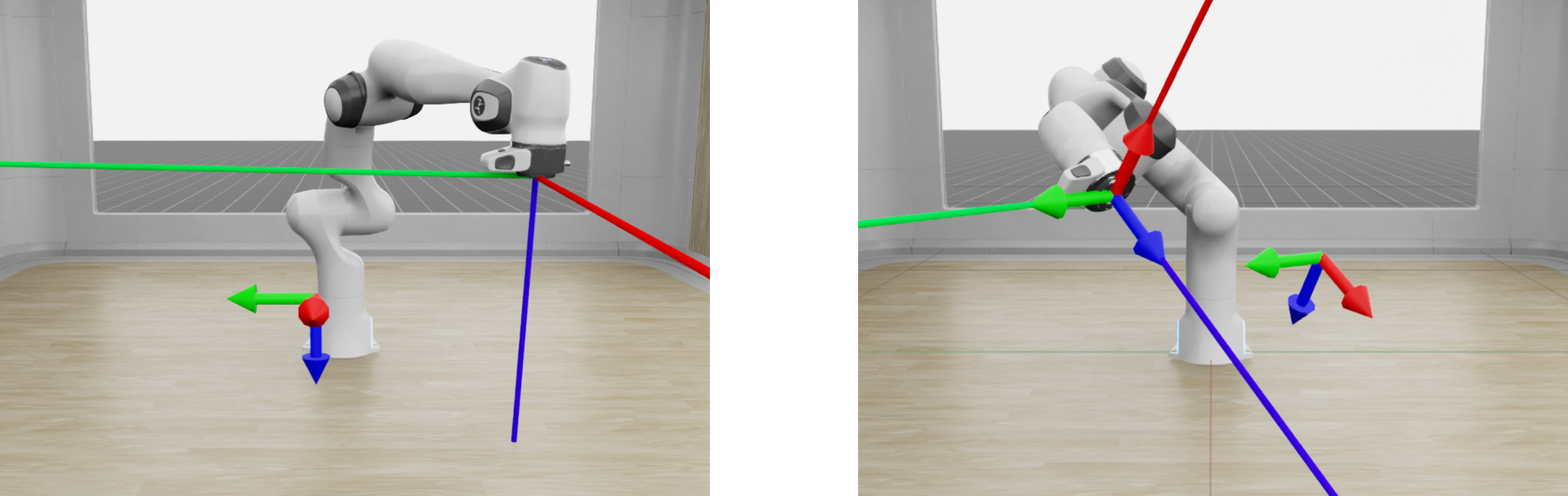}
    \caption{Subset of Calibration Tasks. Left: Position Task (translational motion only). Right: Pose Task (translation and rotational motion).}
    \label{fig:calibration_tasks}
\end{figure}
\vspace{-4mm}

\paragraph{Calibration} Calibration tasks are designed to familiarize users with basic controls. \emph{Position calibration} asks users to place the gripper at randomly spawned targets; \emph{rotation calibration} aligns an attached beam to a target circle; and \emph{pose calibration} combines both position and orientation targets.

\medskip
\paragraph{Evaluation}
Evaluation tasks build on calibration tasks by introducing accuracy and precision measurements along with time constraints. \emph{Position (accuracy)} tasks require reaching position targets under time limits; \emph{rotation (accuracy)} tasks align the beam to disappearing targets; \emph{pose (accuracy)} tasks demand full 6-DoF alignment under time pressure; and \emph{beam (precision)} tasks trace line trajectories of decreasing thickness.

Performance metrics from these tasks inform user proficiency and data quality. Analysis can also reveal which aspects of each input modality are most responsible for errors.

\subsection{\textbf{Performance Metrics}}
\label{subsec:metrics}

We developed multiple metrics to quantify the quality of demonstrations, helping to evaluate the efficacy of input devices, training curricula, ergonomics, and data utility:

\textbf{Task Completion Time} reflects the amount of time taken to complete a task successfully. Shorter times generally indicate higher efficiency.

\textbf{Network Latency} conveys information that helps to understand the implications of network delays on teleoperation. It is measured by calculating the time delay from when a message is sent from the client to when it is received by the server, synchronized with a server clock.

\textbf{Trajectory Path Length} measures the total translational and rotational distance traveled during a demonstration. We define two path length metrics:

\textit{a) Total Translational Distance: }
Let $\{\mathbf{p}_0, \mathbf{p}_1, \dots, \mathbf{p}_{T}\}$ be a sequence of end-effector positions in $\mathbb{R}^3$ measured over the course of a teleoperated episode. The \emph{total translational distance} is the sum of instantaneous translational displacements: $$D_{\text{trans}} \;=\; \sum_{t=0}^{T-1} \|\mathbf{p}_{t+1} - \mathbf{p}_{t}\|_2.$$
A larger value indicates the end-effector traveled a greater distance during teleoperation, suggesting less efficient motion.

\textit{b) Total Rotational Distance: }
Let $\{\mathbf{R}_0, \mathbf{R}_1, \dots, \mathbf{R}_{T}\}$ be a sequence of orientation matrices in $\mathrm{SO}(3)$. For each consecutive pair $(\mathbf{R}_t, \mathbf{R}_{t+1})$, define the relative rotation matrix and the angle of rotation: $$\mathbf{R}_{\text{rel}} \;=\; \mathbf{R}_{t}^{\mathsf{T}} \,\mathbf{R}_{t+1}, \quad \theta_{t} \;=\; \arccos \left(\frac{\mathrm{trace}(\mathbf{R}_{\text{rel}}) - 1}{2}\right).$$
The \emph{total rotational distance} is the sum of these incremental angles: $D_{\text{rot}} \;=\; \sum_{t=0}^{T-1} \theta_{t}.$ A higher total rotation implies more rotational movement throughout the task execution.

\textbf{Motion Jitter} captures the smoothness or abruptness of motion, as defined by local accelerations:

\textit{a) Mean Translational Jitter: }
Define a sequence of translational positions $\{\mathbf{p}_t\}$. Over a sliding window of size $L$, we compute maximal local translational accelerations and average them to obtain the \emph{mean translational jitter}:
$$J_{\text{trans}} \;=\; \frac{1}{N_L}\sum_{w=1}^{N_L} \left(\max_{t \in [w,w+L-2]} a_{t}\right), \quad a_{t} = \frac{v_{t+1} - v_{t}} {\Delta t'_{t}}, $$ where $v_{t} = (\mathbf{p}_{t+1} - \mathbf{p}_{t}) / \Delta t_{t}$ is the discrete velocity, $\Delta t_{t}$ is the time interval between timestamp $t$ and $t+1$, $\Delta t'_{t}$ is the time difference relevant for the velocity interval (e.g., $(\Delta t_t + \Delta t_{t+1})/2$), and $N_L$ is the number of windows. Larger values indicate more abrupt changes in translational speed.

\begin{figure*}[t]
    \centering
    \includegraphics[width=\linewidth]{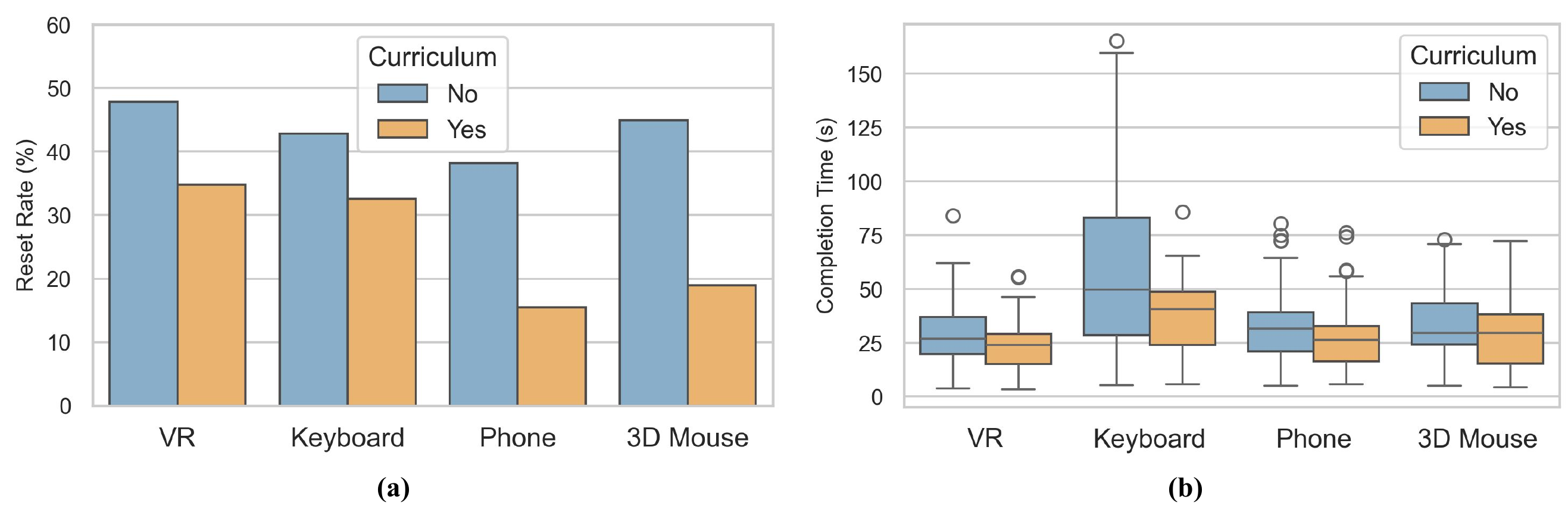}
    \caption{(a) Reset Rate by Device and Curriculum. Across all devices, curriculum training yields a significant decrease in reset rate across tasks, leading to faster and more efficient data collection. (b) Execution Time by Device and Curriculum. Across all devices, curriculum training reduces the mean and standard deviation of execution time, leading to shorter and more consistent demonstrations.}
    \label{fig:reset_bar}
\end{figure*}

\textit{b) Mean Rotational Jitter: }
Similarly, for orientations $\{\mathbf{R}_t\}$, let $\theta_t$ be the incremental rotation angle between $\mathbf{R}_t$ and $\mathbf{R}_{t+1}$. Over a sliding window of size $L$, we compute maximal local angular accelerations and average them: $$J_{\text{rot}} \;=\; \frac{1}{N_L}\sum_{w=1}^{N_L} \left(\max_{t \in [w,w+L-2]} \alpha_{t}\right), \quad \alpha_{t} = \frac{\omega_{t+1} - \omega_{t}}{\Delta t'_{t}}, $$
where $\omega_{t} \;=\; \theta_t / \Delta t_{t}$. Larger rotational jitter values indicate more sudden changes in orientation speed.

\textbf{Communication Loop Jitter} measures simulation and client device stability. We propose two metrics:

\textit{a) Server Loop Jitter: }
During teleoperation, the server processes incoming teleoperation commands at discrete timestamps, which we denote $\{t_0, t_1, \dots, t_{T}\}$. The \emph{server loop jitter} measures the variability in these intervals: $$J_{\text{server}} \;=\; \mathrm{Std}\left(\{t_{t+1} - t_{t}\}_{t=0}^{T-1}\right).$$
where $\mathrm{Std}$ denotes the standard deviation. Lower jitter indicates more consistent server-side loop timing.

\textit{b) Client Loop Jitter: }
On the client side (e.g., user device), we track similarly spaced timestamps $\{\tau_0,\tau_1,\dots,\tau_{T}\}$ at which inputs are sent. The \emph{client loop jitter} is: $$J_{\text{client}} \;=\; \mathrm{Std}\left(\{\tau_{t+1} - \tau_{t}\}_{t=0}^{T-1}\right).$$
Higher client loop jitter suggests variable intervals between sent commands, possibly reflecting inconsistent local processing or network delays affecting the sending rate.

\section{Experiments}
\label{sec:experiments}

We conducted a series of experiments to systematically evaluate \modelName across multiple dimensions, structured around the following research questions:

\noindent \textbf{RQ1}: How does the choice of input device (smartphones, VR, 3D mice, keyboards) affect teleoperation performance, user experience, and data quality?

\noindent \textbf{RQ2}: Does a structured training curriculum improve user proficiency and the quality of collected demonstrations?

\noindent \textbf{RQ3}: Can \modelName scale to support numerous concurrent users while maintaining low end-to-end latency and high simulation control frequency?

\noindent \textbf{RQ4}: What is the cost-efficiency of demonstration collection using \modelName?

\noindent \textbf{RQ5}: Is the data from \modelName useful for training performant behavior cloning policies in simulation?

\noindent \textbf{RQ6}: Is the data from \modelName useful for training performant behavior cloning policies in the real world?

\subsection{\textbf{RQ1 \& RQ2: Input Device Comparison and Curriculum Effectiveness}}
To address RQ1 and RQ2, we recruited 12 participants for an initial user study. All participants provided written informed consent prior to participation. Six participants were randomly assigned to first complete the training curriculum described in Section~\ref{subsec:curriculum}, while the other six served as a control group (no prior training). Each participant used two randomly assigned input devices (chosen from smartphone, VR headset, 3D mouse, keyboard) to perform a set of calibration and manipulation tasks. Participants were instructed to provide five successful demonstrations (per assigned input device) for each of the four manipulation tasks: Three-Piece Assembly, Lift, Mug Cleanup, and Coffee (see \cite{mimicgen} for task details). We collected a set of metrics that include task completion time, total path length, translational and rotational jitter, and task reset rates (Section~\ref{subsec:metrics}). Subjective feedback was collected using NASA-TLX surveys and Likert scale questionnaires focusing on ease of use, comfort, and perceived accuracy. A separate study with six additional participants compared dual-smartphone versus VR control for bimanual tasks.

\textit{(RQ1: Device Comparison)} Performance varied significantly across devices (Table~\ref{tab:device_summary_metrics}). Smartphones and VR headsets generally yielded better objective metrics, including shorter completion times and smoother trajectories (lower jitter) when compared with keyboards and 3D mice. Although keyboards and 3D mice yielded shorter absolute path lengths, this stems from the way their inputs are received. VR controllers and phones transmit continuous pose updates, where even small movements accumulate into longer trajectories, whereas 3D mice support direct velocity control and keyboards restrict control to discrete steps. Nonetheless, the pose evaluation task (Table~\ref{tab:device_error_results}) showed that smartphones achieved the lowest position and rotation errors, demonstrating their effectiveness for capturing high-quality, precise demonstrations. Subjective feedback corroborates these results, with users rating smartphones and VR higher on willingness to use again and subjective comfort. Note that these tasks were conducted in MuJoCo-based environments, so the translational and rotational metrics in Table~\ref{tab:device_summary_metrics} and Table~\ref{tab:device_error_results} are unitless and should be interpreted only in relative terms.

\begin{figure*}[t]
  \centering
  \includegraphics[width=1.0\linewidth]{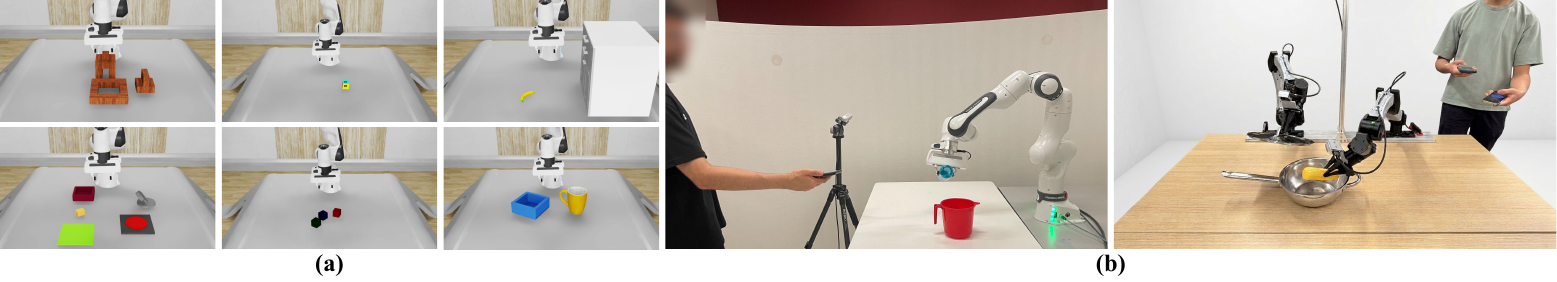}
  \caption{(a) Visualization of Isaac Lab tasks in the pilot dataset. Arrangement of tasks left-to-right, top-to-bottom: Assembly, Lift, Cleanup, Kitchen, Stack, Pour. (b) \modelName can be used to control physical (single-arm and bimanual) robots. A real-world recreation of the pour task and a corn cooking task are shown.}
  \label{fig:banner_tasks_teleop}
\end{figure*}

\begin{table*}[!t]
  \centering
  \begin{minipage}{0.70\textwidth}
    \centering
    \resizebox{\linewidth}{!}{
      \begin{tabular}{lcllll}
        \toprule
        \textbf{Metric} & & \textbf{Smartphone} & \textbf{VR Headset} & \textbf{3D Mouse} & \textbf{Keyboard} \\
        \midrule
        Avg. Completion Time (s) & ($\downarrow$) & $30.00$\footnotesize{$\pm16.97$} & $\textbf{25.60}$\footnotesize{$\pm13.91$} & $31.14$\footnotesize{$\pm17.12$} & $46.49$\footnotesize{$\pm32.74$} \\
        \rowcolor[HTML]{EFEFEF}
        Avg. Translational Path Length & ($\downarrow$) & $2.47$\footnotesize{$\pm1.57$} & $2.30$\footnotesize{$\pm1.29$} & $\textbf{1.95}$\footnotesize{$\pm1.02$} & $2.00$\footnotesize{$\pm1.09$} \\
        Avg. Rotational Path Length & ($\downarrow$) & $4.23$\footnotesize{$\pm2.74$} & $4.18$\footnotesize{$\pm2.91$} & $\textbf{1.63}$\footnotesize{$\pm1.17$} & $2.03$\footnotesize{$\pm1.48$} \\
        \rowcolor[HTML]{EFEFEF}
        Avg. Translational Jitter & ($\downarrow$) & $\textbf{0.24}$\footnotesize{$\pm0.09$} & $0.43$\footnotesize{$\pm0.18$} & $0.35$\footnotesize{$\pm0.18$} & $0.65$\footnotesize{$\pm0.26$} \\
        Avg. Rotational Jitter & ($\downarrow$) & $\textbf{0.37}$\footnotesize{$\pm0.14$} & $0.81$\footnotesize{$\pm0.39$} & $0.44$\footnotesize{$\pm0.21$} & $0.64$\footnotesize{$\pm0.21$} \\
        \rowcolor[HTML]{EFEFEF}
        Reset Rate (\%) & ($\downarrow$) & \textbf{28.57} & 42.03 & 34.43 & 38.14 \\ 
        Willing to Use Again (1-5) & ($\uparrow$) & $\textbf{4.33}$\footnotesize{$\pm0.52$} & $4.17$\footnotesize{$\pm1.17$} & $3.83$\footnotesize{$\pm1.33$} & $3.17$\footnotesize{$\pm1.47$} \\
        \rowcolor[HTML]{EFEFEF}
        Subjective Comfort (1-5) & ($\uparrow$) & $\textbf{4.50}$\footnotesize{$\pm0.55$} & $4.33$\footnotesize{$\pm1.03$} & $3.67$\footnotesize{$\pm1.21$} & $3.33$\footnotesize{$\pm1.03$} \\
        \bottomrule
      \end{tabular}
    }
  \end{minipage}\hfill
  \begin{minipage}{0.27\textwidth}
    \captionof{table}{\textbf{Summary of Key Performance Metrics Across Input Devices in User Study (Mean $\pm$ Std. Dev.).} We observe that the use of \modelName\ with a smartphone generally yields improved quantitative and qualitative metrics. Note that path length and jitter are unitless.}
    \label{tab:device_summary_metrics}
  \end{minipage}
\end{table*}

\begin{table}[t]
    \centering
    \caption{\textbf{Pose Evaluation Task Average Positional and Rotational Error Across Input Devices.} Smartphones yielded significantly lower errors than the other input modalities.}
    \label{tab:device_error_results}
        \begin{tabular}{lcc}
            \toprule
            \textbf{Device} & \textbf{Position Error} & \textbf{Rotation Error} \\
            \midrule
            \rowcolor[HTML]{EFEFEF}
            Phone & $\textbf{0.13}$\footnotesize{$\pm0.06$} & $\textbf{0.29}$\footnotesize{$\pm0.17$} \\
            VR Headset & $0.20$\footnotesize{$\pm0.20$} & $0.51$\footnotesize{$\pm0.39$} \\
            \rowcolor[HTML]{EFEFEF}
            3D Mouse & $0.25$\footnotesize{$\pm0.13$} & $1.36$\footnotesize{$\pm0.53$} \\
            Keyboard & $0.16$\footnotesize{$\pm0.06$} & $0.87$\footnotesize{$\pm0.28$} \\
            \bottomrule
        \end{tabular}
\end{table}

For bimanual tasks, dual-smartphone control was identified as a low-cost alternative to VR demonstrations. Models were trained on a small dataset (60 demonstrations) of dual-smartphone and VR data. With BC-RNN and BC-Transformer, we obtained success rates of $22\%$ and $26\%$, respectively. Success rates were calculated based on $100$ rollouts. These results validate the use of \modelName for bimanual teleoperation.

\textit{(RQ2: Curriculum Effectiveness)} The training curriculum had a demonstrably positive impact. Participants who underwent training exhibited significantly lower reset rates (Figure~\ref{fig:reset_bar}a) and reduced mean execution times across all devices (Figure~\ref{fig:reset_bar}b). This suggests the curriculum effectively onboarded users, improving data collection speed and quality by reducing errors and increasing device familiarity.

To assess statistical significance, we compared groups with and without the training curriculum using independent tests. For completion time, a one-sided t-test tested whether participants who received the curriculum executed downstream tasks in a shorter amount of time. The result yielded a test statistic of $-4.26$ and a p-value less than $0.0001$, indicating statistical significance. For reset rate, we conducted a two-sample proportion test to determine whether the training curriculum reduced the frequency of resets. This test produced a statistic of $-4.88$ and a p-value of $1 \times 10^{-6}$, also significant. Together, these results demonstrate that a structured training curriculum substantially improves both the efficiency and reliability of trajectories collected by novice teleoperators.

\subsection{\textbf{RQ3 \& RQ4: Scalability, Latency, and Cost Efficiency}}

Guided by the findings that smartphones and VR offer superior performance and user experience, we evaluated the scalability of \modelName for crowdsourced data collection.

We deployed \modelName on Google Cloud Platform (GCP) following the architecture in Figure~\ref{fig:method}. We varied the number of concurrent teleoperators connecting to a single GPU instance running vectorized Isaac Lab environments from one user up to eight to observe scaling effects. We measured:

\textit{a) Average latency}: Time from user input action to client session service receiving the action.

\textit{b) Simulation control loop time}: Time taken by the simulation environment to process user commands and update its state.

\textit{c) Resource utilization}: RAM and VRAM usage.

\textit{d) Cost per 1,000 demonstrations}: Estimated cost for 1,000 demonstrations based on cloud pricing.

\begin{table*}[!t]
  \centering
  \begin{minipage}{0.70\textwidth}
    \centering
    \resizebox{\linewidth}{!}{
      \begin{tabular}{ccccc}
        \toprule
        \textbf{\# Users} & \textbf{Avg. Latency (ms)} & \textbf{Med. Sim Step (ms)} & \textbf{Peak VRAM (GB)} & \textbf{Peak RAM (GB)} \\
        \midrule
        $1$ & $1.70 \pm 4.95$ & $45.68 \pm 0.09$ & $3.40 \pm 0.00$ & $5.17 \pm 0.00$ \\
        \rowcolor[HTML]{EFEFEF}
        $2$ & $6.16 \pm 4.97$ & $51.08 \pm 0.02$ & $3.41 \pm 0.00$ & $5.37 \pm 0.00$ \\
        $4$ & $4.79 \pm 4.82$ & $60.06 \pm 0.08$ & $3.55 \pm 0.00$ & $5.70 \pm 0.06$ \\
        \rowcolor[HTML]{EFEFEF}
        $8$ & $7.08 \pm 4.93$ & $79.31 \pm 0.04$ & $3.88 \pm 0.00$ & $6.66 \pm 0.02$ \\
        \bottomrule
      \end{tabular}
    }
  \end{minipage}\hfill
  \begin{minipage}{0.27\textwidth}
    \captionof{table}{\textbf{System Scaling}. System Performance vs. Number of Concurrent Users per GPU (NVIDIA T4). As the number of concurrent clients on a single GPU increases, latency and memory utilization grow sublinearly.}
    \label{tab:scaling_performance}
  \end{minipage}
\end{table*}

\textit{(RQ3: Scalability \& Performance)} Our architecture demonstrated effective scaling, sustaining multiple concurrent users while maintaining interactive performance. Although teleoperation remained feasible when scaled to eight concurrent users, we decided to limit active sessions on one GPU to four to ensure the best user experience (Table~\ref{tab:scaling_performance}). Memory utilization marginally increased when scaling users, demonstrating that CPU-bound processes such as WebSocket communication and WebRTC streaming impose minimal overhead on overall system performance. Additionally, we conducted system load testing by simulating \textit{256 concurrent clients} distributed over \textit{8 GPUs}. \modelName sustained this workload in a distributed fashion with a median simulation-step latency of 186.7 ms. Importantly, our results show that this latency does not scale linearly with client count, indicating that more powerful GPUs may deliver \textit{superlinear scaling} improvements.

\textit{(RQ4: Cost Efficiency)} Cloud deployment offers flexibility and scalability, but without effective system resource utilization, scaling can be expensive. By serving multiple clients on a single GPU, \modelName substantially cuts these costs. On an NVIDIA T4 instance, priced at \$0.92 per hour (estimated GCP cost), assuming each user completes 120 demonstrations in that hour, the cost for 1,000 demonstrations in a non-vectorized setting would be \$7.67. With \modelName, this cost drops to just \$1.92. Under these assumptions, this equates to a near \textit{4x reduction} in data collection expenses on entry‐level hardware. \modelName can flexibly adjust concurrency to achieve an optimal cost–performance balance, scaling to support 12 concurrent users on a high-end GPU like the NVIDIA RTX 6000.

\subsection{\textbf{RQ5: Data Validation via Behavior Cloning}}

\noindent Finally, to demonstrate the practicality of crowdsourced data, we collected a large pilot dataset using \modelName, leveraging insights from our initial studies. We then used our pilot dataset to train several imitation learning policies.

\begin{table}[t]
    \centering
    \captionof{table}{\textbf{\modelName Pilot Dataset Statistics}}
    \label{tab:bc_stats}

    \begin{tabular}{lccc}
        \toprule
        Task & Demonstrations & Hours \\
        \midrule
        Lift & 1,294 & 1.99 \\
        \rowcolor[HTML]{EFEFEF}
        Pour & 1,026 & 4.92 \\
        Stack & 1,112 & 5.77 \\
        \rowcolor[HTML]{EFEFEF}
        Cleanup & 1,023 & 6.86 \\
        Assembly & 1,007 & 7.14 \\
        \rowcolor[HTML]{EFEFEF}
        Kitchen & 1,284 & 17.16 \\
        User Study & 764 & 6.77 \\
        \midrule
        \textbf{Total} & 7,510 & 50.61 \\
        \bottomrule
    \end{tabular}
\end{table}

We crowdsourced the collection of 6,746 human demonstrations using only smartphones across several benchmark tasks in Isaac Lab (Table~\ref{tab:bc_stats}, Figure~\ref{fig:banner_tasks_teleop}a). These tasks included four Isaac Lab environments created from scratch and two modified stock environments. Data quality was maintained by filtering based on performance metrics (Section~\ref{subsec:metrics}), selecting demonstrations within the 50th percentile for total path length to remove suboptimal trajectories. We trained standard behavior cloning algorithms (BC-RNN, BC-Transformer \cite{robomimic}) and state-of-the-art methods like Action Chunking with Transformers (ACT) \cite{aloha} and Diffusion Policy (DP) \cite{chi2023diffusion} on this curated dataset. Policy performance was evaluated based on task success rates over $50$ rollouts per task.

\begin{table}[t]
    \centering
    \captionof{table}{\textbf{BC Results Per Task.} Data collected with \modelName is capable of achieving a variety of tasks using SOTA algorithms.}
    \label{tab:advanced_bc_results}

    \begin{tabular}{lcccc}
        \toprule
        \textbf{Task} & \textbf{BC-RNN} & \textbf{BC-TF} & \textbf{ACT} & \textbf{DP} \\
        \midrule
        Lift & \textbf{1.00} & 0.84 & 0.88 & \textbf{1.00} \\
        \rowcolor[HTML]{EFEFEF}
        Pour & \textbf{0.68} & 0.36 & 0.36 & 0.54 \\
        Stack & 0.00 & 0.00 & \textbf{0.60} & 0.58 \\
        \rowcolor[HTML]{EFEFEF}
        Cleanup & 0.72 & 0.20 & 0.92 & \textbf{0.94} \\
        Assembly & 0.36 & 0.10 & 0.32 & \textbf{0.50} \\
        \rowcolor[HTML]{EFEFEF}
        Kitchen & 0.04 & 0.02 & 0.10 & \textbf{0.12} \\
        \bottomrule
    \end{tabular}
\end{table}

The policies trained on the \modelName-collected data achieved high success rates on the majority of tasks from our task suite (Table~\ref{tab:advanced_bc_results}), confirming the effectiveness of metric-based filtering and overall quality of crowdsourced demonstrations. Note that BC-RNN and BC-TF were trained on additional low-dimensional observations, specifically task-relevant object poses. Capturing additional data or observations (such as another camera view) may improve these results. Nonetheless, these results confirm that \modelName can produce datasets effective for training robot manipulation policies, and that data collected via accessible devices like smartphones captures sufficient fidelity and diversity.

\subsection{\textbf{RQ6: Real-Robot Compatibility}}

In addition to our simulation environments, we validated our smartphone teleoperation pipeline on multiple platforms, including a Franka Panda arm and bimanual YAM arms (Figure~\ref{fig:banner_tasks_teleop}b). We configured our system to connect the mobile app directly to a server running on the robot’s host machine over our local network, enabling low-latency, real-time control. To ensure safe deployment on hardware, we incorporate safety mechanisms into the teleoperation pipeline. For example, the smartphone automatically disables robot control if its velocity exceeds a threshold, preventing accidental motions (e.g., if the user drops the phone). We validated our Franka setup by collecting 98 expert demonstrations on the standard Lift task and then training a BC-RNN policy on this data, achieving a 52\% success rate over 25 rollouts. Although lower than the simulation results, this confirms the ability to use \modelName with real hardware. We expect performance to improve with more real-world data collection on \modelName, which we leave to future work. Nonetheless, these results confirm that our phone-based interface can be applied to real-world hardware with minimal setup.

\section{Conclusion}
\label{sec:conclusion}

By lowering the barrier to entry for remote teleoperation, \modelName aims to democratize large-scale dataset creation for imitation learning. Through a comprehensive user study, we demonstrate that integrating accessible devices, user-friendly interfaces, and robust networking infrastructure can significantly improve the quality and efficiency of data collection. We also establish core metrics to rank operators, assess server performance, evaluate a device's effectiveness in producing demonstrations, and identify the most reliable and high-quality demos. Our infrastructure enabled the crowdsourcing of a high-quality pilot dataset with over 7,500 trajectories, which we used to successfully train imitation learning models. While platforms like \modelName address the critical bottleneck of operator availability and data volume, they highlight the emergence of a new potential bottleneck: \textit{the creation of diverse, high-quality simulation environments}. As collecting demonstrations at scale becomes easier, achieving task diversity and reliable generalization will require a parallel effort in scaling task design. Ultimately, future progress in robot learning will depend on both accessible data collection platforms like \modelName as well as consistent scaling of novel, complex environments.

\section*{Acknowledgments}

We thank all of our user study participants for providing valuable data and feedback to improve our platform.

\bibliographystyle{IEEEtran}
\bibliography{bib}

\newpage
\appendices 

\section{Input Device Details}
\label{app:input_devices}

\modelName is compatible with a diverse set of input devices:

\subsection{\textbf{Smartphones}}
Smartphones provide accurate 6-DoF pose tracking by utilizing existing AR frameworks (ARCore for Android and ARKit for iOS). Our cross-platform mobile application captures both translational and rotational motion, transmitting the change in pose at each timestep.

\subsection{\textbf{Virtual Reality Headsets}}
\modelName supports the Meta Quest 3 for immersive VR input. Our Unity application runs natively on the headset, allowing users to teleoperate by wearing the headset and using the handheld Quest controllers. The app transmits the pose of the handheld controllers in the world frame using Meta's built-in tracking system.

\subsection{\textbf{Keyboards}}
Keyboard control is implemented by mapping specific keys to translational and rotational movements, while additional keys manage reset functions and the robot's gripper state. This setup provides an intuitive and straightforward method of control for the robot's movements and actions.

\subsection{\textbf{3D Mice}}
The 3Dconnexion SpaceMouse offers 6-DoF control through its ergonomic joystick and buttons, facilitating precise and fluid manipulation. Its hardware strengths, in conjunction with intuitive switching between translation and rotation modes, offer smooth trajectories and enhance the teleoperation experience.


\section{Task Descriptions}
\label{app:tasks}

The user studies and data collection involved several manipulation tasks designed to evaluate teleoperation performance and collect diverse demonstration data.

\subsection{\textbf{User Study Tasks}}
\label{app:user-study-tasks}

\begin{figure}[!ht]
  \centering
  \includegraphics[width=\linewidth]{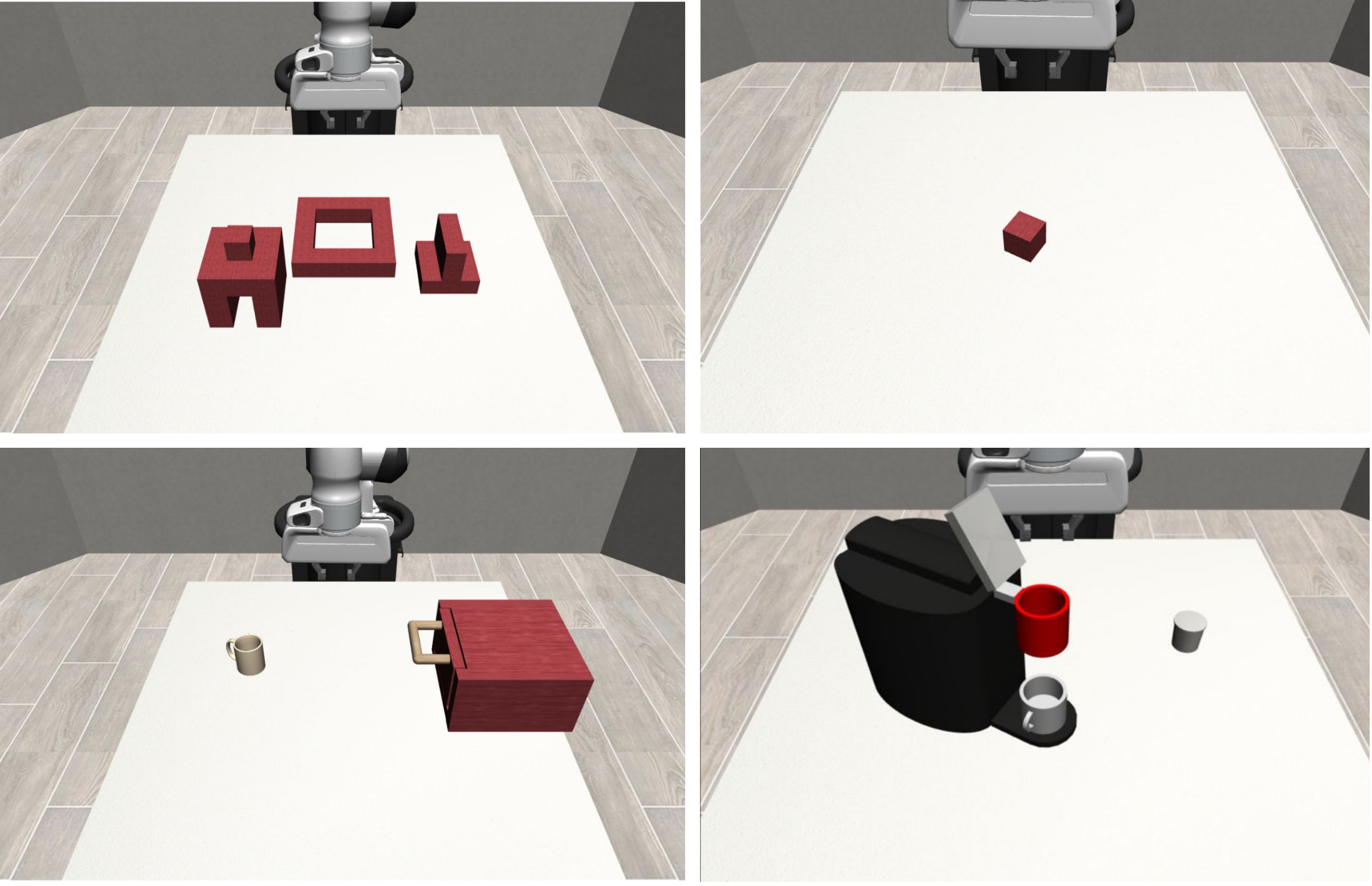}
  \caption{Visualization of the primary user study tasks. Top row (left to right): Three Piece Assembly, Lift. Bottom row (left to right): Mug Cleanup, Coffee.}
  \label{fig:tasks_app}
\end{figure}

Our user study used four MimicGen \cite{mimicgen} tasks: Lift, Three Piece Assembly, Mug Cleanup, and Coffee. Screenshots of the four tasks are shown in Figure~\ref{fig:tasks_app}.

\begin{figure}[!ht]
  \centering
  \includegraphics[width=\linewidth]{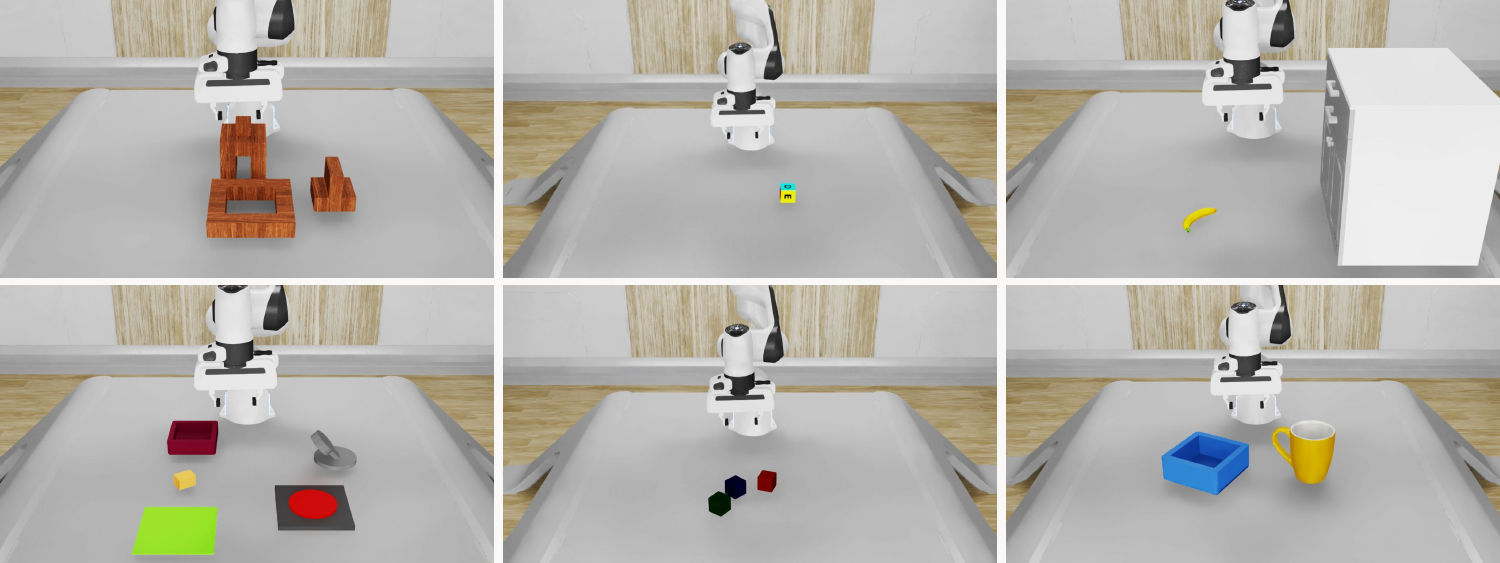}
  \caption{Visualization of the tasks in the pilot dataset. Top row (left to right): Assembly, Lift, Cleanup. Bottom row (left to right): Kitchen, Stack, Pour.}
  \label{fig:dataset_tasks}
\end{figure}

\subsection{\textbf{Dataset Tasks}}
\label{app:dataset-tasks}

To create a dataset that captures a wide range of manipulator use cases, we collected data for six environments in Isaac Lab. Two environments (Lift, Stack) were slightly modified from the default Isaac Lab environments. We recreated three MimicGen tasks (Assembly, Cleanup, Kitchen) in Isaac Lab and designed a custom environment as well (Pour). Brief descriptions of each task are written below, and screenshots are shown in Figure~\ref{fig:dataset_tasks}.

\begin{itemize}
    \item \textbf{Lift}: Grasp the cube and lift it above the table.
    \item \textbf{Stack}: Place the red cube on the blue cube. Then, place the green cube over the red cube.
    \item \textbf{Assembly}: Place the inverted-T shape object inside the base. Then, place the cap object on top.
    \item \textbf{Cleanup}: Place the banana in the top drawer.
    \item \textbf{Pour}: Use the mug to pour the sphere into the bin.
    \item \textbf{Kitchen}: Place the bread in the pot, place the pot on the stove, and turn on the switch. Then, place the pot in the green region and turn off the switch.
\end{itemize}

\section{User Study Details}
\label{app:user_study}

\subsection{\textbf{Experimental Procedure}}
\begin{enumerate}
    \item \textbf{Recruitment}: A total of $18$ consenting participants were recruited.
    \item \textbf{Training Condition}: Half of the participants were randomly assigned to the training group and completed the curriculum (Section~\ref{subsec:curriculum}) before the main tasks. The other half formed the control group (no prior training).
    \item \textbf{Input Device Assignment}: Participants were randomly assigned two input devices from: smartphone, virtual reality (VR) headset, keyboard, and 3D mouse. The order of devices was also randomized. Participants prone to motion sickness were excluded from VR.
    \item \textbf{Task Performance}: Each participant used their assigned devices to perform four distinct manipulation tasks (Lift, TPA, MC, Coffee - see Figure \ref{fig:tasks_app}), providing five successful demonstrations per task per device.
    \item \textbf{Data Collection:} Demonstration data (trajectory, timings, resets) and system metrics (latency, jitter) were collected during the tasks.
    \item \textbf{Survey Administration}: After completing all tasks with one device, participants completed the Likert-scale questionnaire and the NASA-TLX survey. They were allowed to revisit responses after using the second device.
    \item \textbf{Bimanual Control Assessment}: A separate group of 6 participants evaluated bimanual control for the Two Arm Lift task. Participants were randomly assigned to use either dual smartphones or a VR system first, then switched. Participants prone to motion sickness were excluded from VR.
\end{enumerate}

\begin{table*}[!ht]
    \centering
    \begin{minipage}[t]{0.62\textwidth}
        \vspace{0pt}
        \centering
        \caption{Behavior Cloning (BC) Model Success Rates (User Study Tasks)}
        \label{tab:bc_model_success_rates_app}
        \begin{tabular}{lcccc}
            \toprule
            \textbf{Model} & \textbf{Lift} & \textbf{TPA} & \textbf{MC} & \textbf{Coffee} \\
            \midrule
            \textbf{BC-RNN} & $1.00 \pm 0.00$ & $0.00 \pm 0.01$ & $0.61 \pm 0.01$ & $0.49 \pm 0.03$ \\
            \textbf{BC-TF}  & $0.90 \pm 0.02$ & $0.03 \pm 0.01$ & $0.64 \pm 0.08$ & $0.36 \pm 0.21$ \\
            \bottomrule
        \end{tabular}
        \vspace{1mm}
        \begin{flushleft}
        {\footnotesize Note: Each model trained on 3 seeds. Success rates are Mean $\pm$ Std. Dev. over 100 rollouts per seed. TPA: Three Piece Assembly, MC: Mug Cleanup, BC-TF: BC-Transformer.}
        \end{flushleft}
    \end{minipage}\hfill
    \begin{minipage}[t]{0.36\textwidth}
        \centering
        \caption{Behavior Cloning (BC) Model Success Rates (Bimanual Task)}
        \label{tab:bimanual_success_rates_app}
        \begin{tabular}{lc}
            \toprule
            \textbf{Model} & \textbf{Two Arm Lift} \\
            \midrule
            \textbf{BC-RNN} & $0.22$ \\
            \textbf{BC-Transformer} & $0.26$ \\
            \bottomrule
        \end{tabular}
        \vspace{1mm}
        \begin{flushleft}
        {\footnotesize Note: Trained on bimanual user study data. Success rate over 100 rollouts.}
        \end{flushleft}
    \end{minipage}
\end{table*}

\subsection{\textbf{User Study Dataset Statistics}}
The user study conducted in this work involved collecting 764 demonstrations, totaling 6.77 successful hours, among 18 participants. Table \ref{tab:user_study_task_statistics} describes the number of demonstrations and hours of data per task.

\begin{table}[ht]
    \centering
    \caption{User Study Dataset Statistics}
    \label{tab:user_study_task_statistics}

    \begin{tabular}{lccc}
        \toprule
        Task & Demonstrations & Hours \\
        \midrule
        Lift & 204 & 0.55 \\
        \rowcolor[HTML]{EFEFEF}
        Three Piece Assembly & 120 & 1.56 \\
        Mug Cleanup & 120 & 1.23 \\
        \rowcolor[HTML]{EFEFEF}
        Coffee & 204 & 1.98 \\
        Two Arm Lift & 60 & 0.23 \\
        \rowcolor[HTML]{EFEFEF}
        Two Arm Transport & 56 & 1.22 \\
        \midrule
        \textbf{Total} & 764 & 6.77 \\
        \bottomrule
    \end{tabular}
\end{table}

\subsection{\textbf{Behavior Cloning Results (User Study Data)}}
The demonstration data collected during the user study were used to train baseline BC models. Table \ref{tab:bc_model_success_rates_app} illustrates the success rates across the four user study tasks with BC-RNN and BC-Transformer. Table \ref{tab:bimanual_success_rates_app} illustrates the success rates on a bimanual user study task with BC-RNN and BC-Transformer.

\subsection{\textbf{Survey Instruments}}

\subsubsection{\textbf{NASA Task Load Index (NASA-TLX) Metrics}}
Participants rated the following aspects of the teleoperation experience on a scale from 1 (low) to 20 (high):
\begin{enumerate}[leftmargin=2em, label={\arabic*.}, topsep=2pt]
    \item \textbf{Mental Demand}: Level of mental and perceptual activity required (thinking, deciding, calculating, remembering, looking, searching).
    \item \textbf{Physical Demand}: Amount of physical activity required (pushing, pulling, turning, controlling, activating).
    \item \textbf{Temporal Demand}: Amount of time pressure felt due to the rate or pace at which the tasks or task elements occurred.
    \item \textbf{Performance}: How successful the participant felt they were in accomplishing the goals of the task set by the experimenter (their own performance).
    \item \textbf{Effort}: How hard the participant had to work (mentally and physically) to accomplish their level of performance.
    \item \textbf{Frustration}: How insecure, discouraged, irritated, stressed, and annoyed versus secure, gratified, content, relaxed, and complacent the participant felt during the task.
\end{enumerate}

\subsubsection{\textbf{Likert-Scale Questions}}
Participants responded to the following statements on a Likert scale from 1 (strongly disagree) to 5 (strongly agree):
\begin{enumerate}[leftmargin=2em, label={\arabic*.}, topsep=2pt]
    \item I found it easy to control the robot with the device I used.
    \item The interface felt intuitive.
    \item I felt comfortable and confident throughout the task.
    \item I would be willing to use this system again in the future.
    \item The tasks seemed appropriate for this type of interface.
    \item My input device responded accurately to my actions.
\end{enumerate}

\subsubsection{\textbf{Open-Ended Questions}}
Participants provided qualitative feedback on their experience by answering:
\begin{enumerate}[leftmargin=2em, label={\arabic*.}, topsep=2pt]
    \item What did you like most about controlling the robot with this device?
    \item What did you find most difficult or frustrating?
\end{enumerate}

\subsection{\textbf{Additional User Study Figures}}

This section contains figures presenting results from the user study surveys and specific evaluation tasks.

\begin{figure}[h!] 
    \centering
    \includegraphics[width=\linewidth]{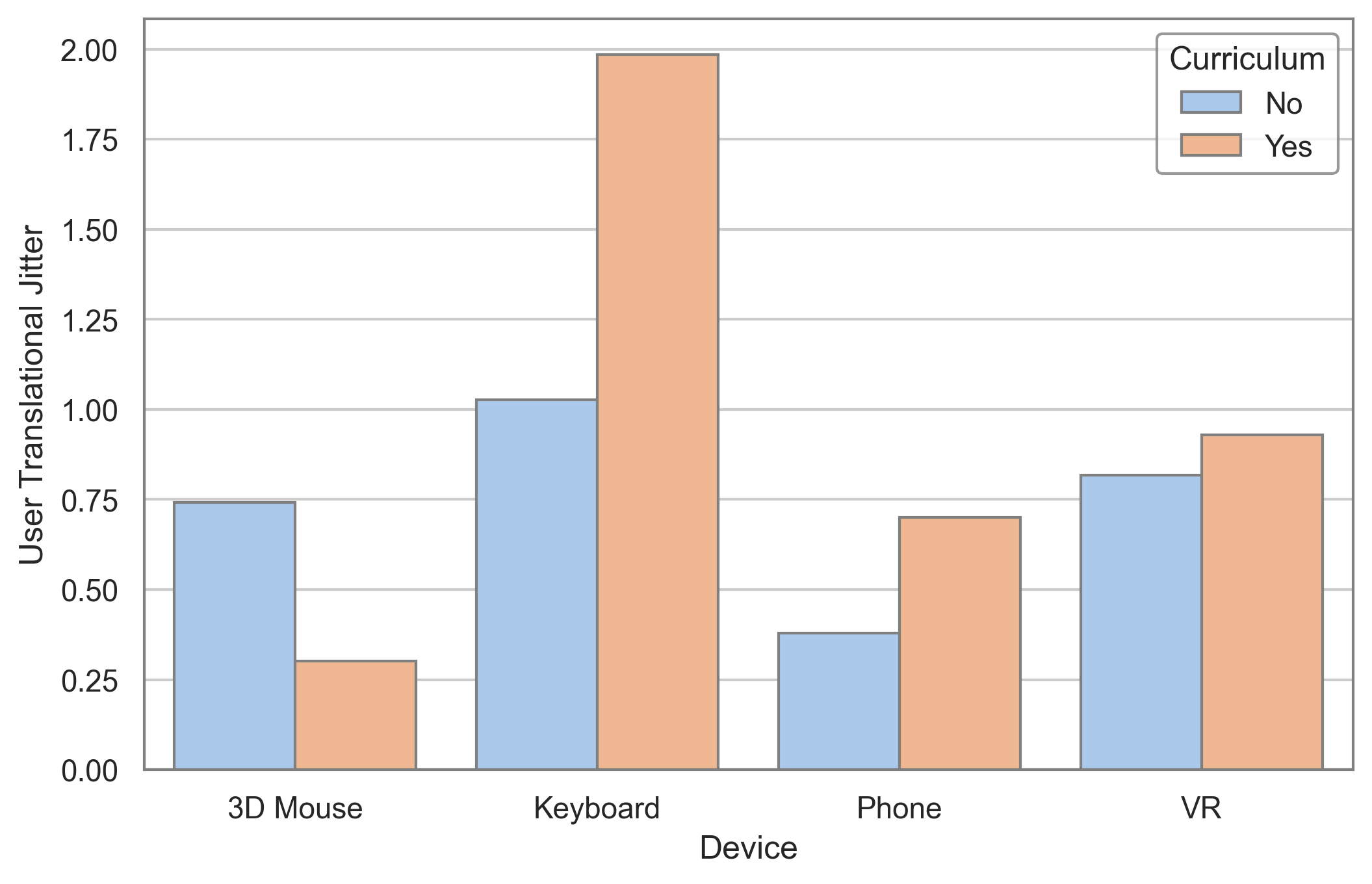}
    \caption{Mean Translational Jitter by device and curriculum condition during the \textit{Position Evaluation Task} (Lower is better). Error bars indicate standard deviation.}
    \label{fig:user_translation_jitter_app} 
\end{figure}

\begin{figure*}[!t]
    \centering
    \includegraphics[width=\linewidth]{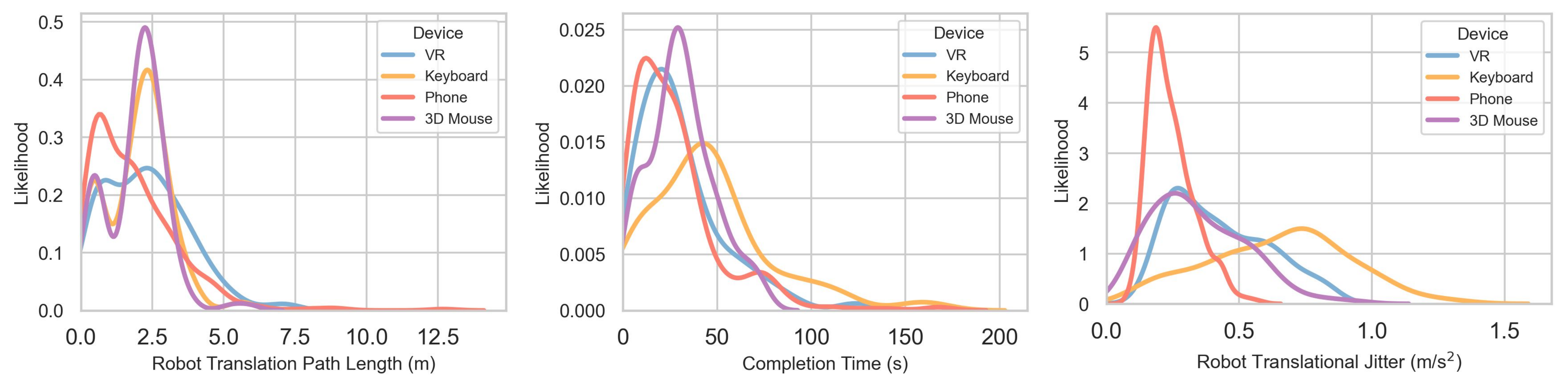}
    \caption{Path length, completion time, and translational jitter per device. These plots reveal performance differences across devices, with smartphones and VR headsets generally yielding shorter path lengths and faster completion times.}
    \label{fig:device_pathlength}
\end{figure*}

\begin{figure}[!t]
    \centering
    \includegraphics[width=\linewidth]{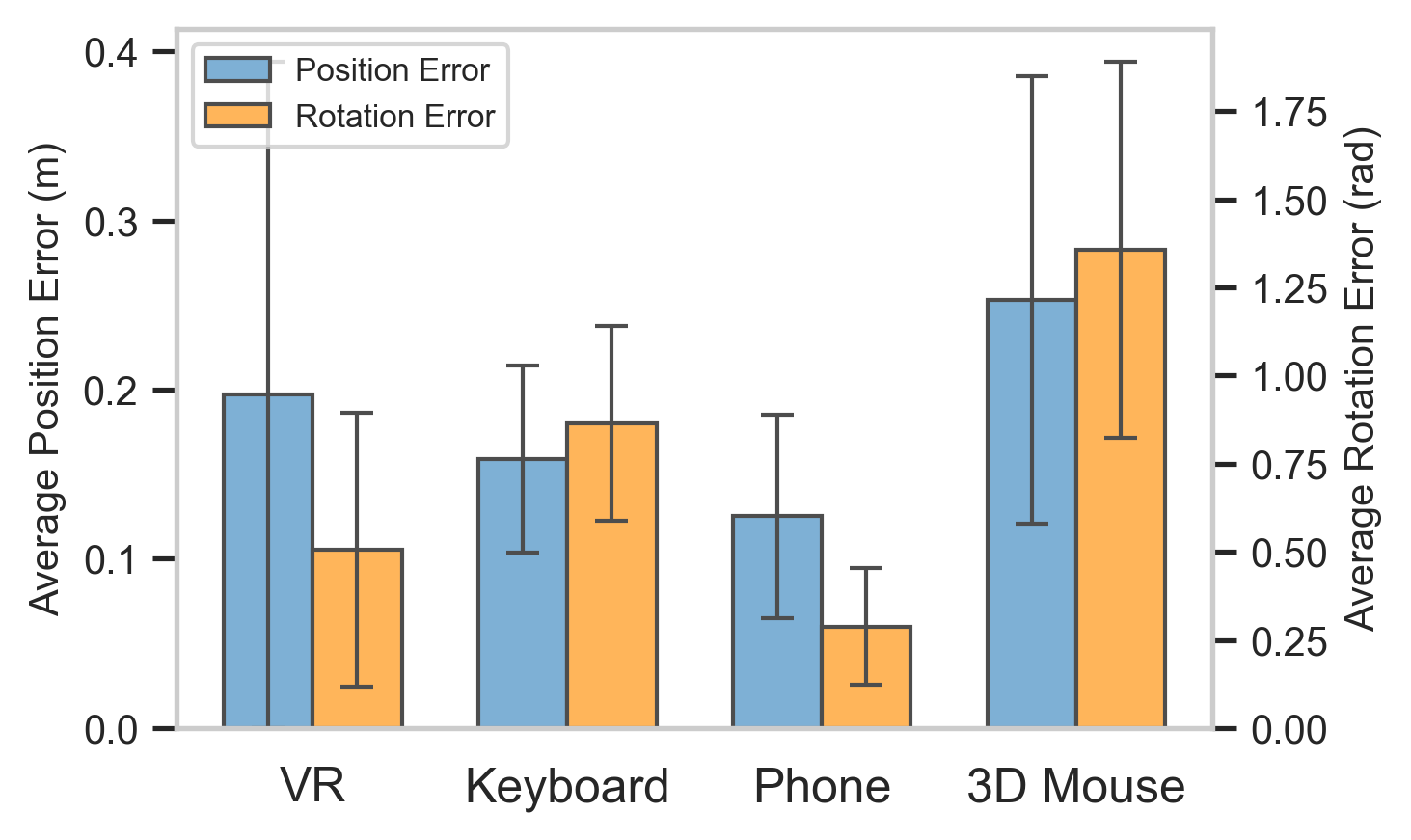}
    \caption{Pose evaluation task's average position and rotation error by device. The smartphone input modality was shown to have a significantly lower position and rotation error than the other input modalities.}
    \label{fig:device_error}
\end{figure}

\begin{figure}[!t] 
    \centering
    \includegraphics[width=\linewidth]{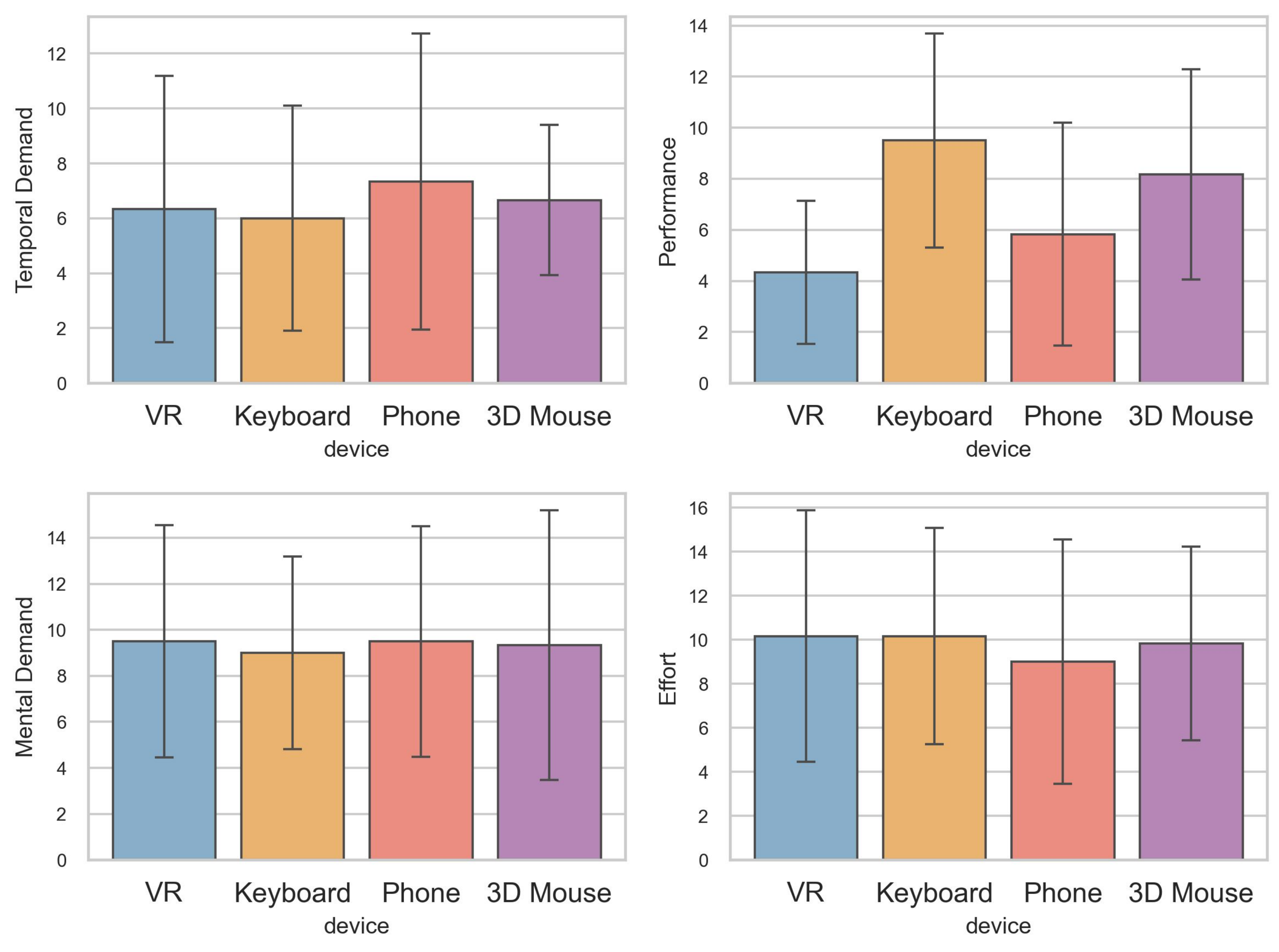}
    \caption{Additional NASA-TLX results (Mean scores per device). Lower scores generally indicate lower perceived workload (except for Performance, where higher is better). Error bars show standard deviation.}
    \label{fig:nasa-tlx-supp_app} 
\end{figure}

\begin{figure}[!t] 
    \centering
    \includegraphics[width=\linewidth]{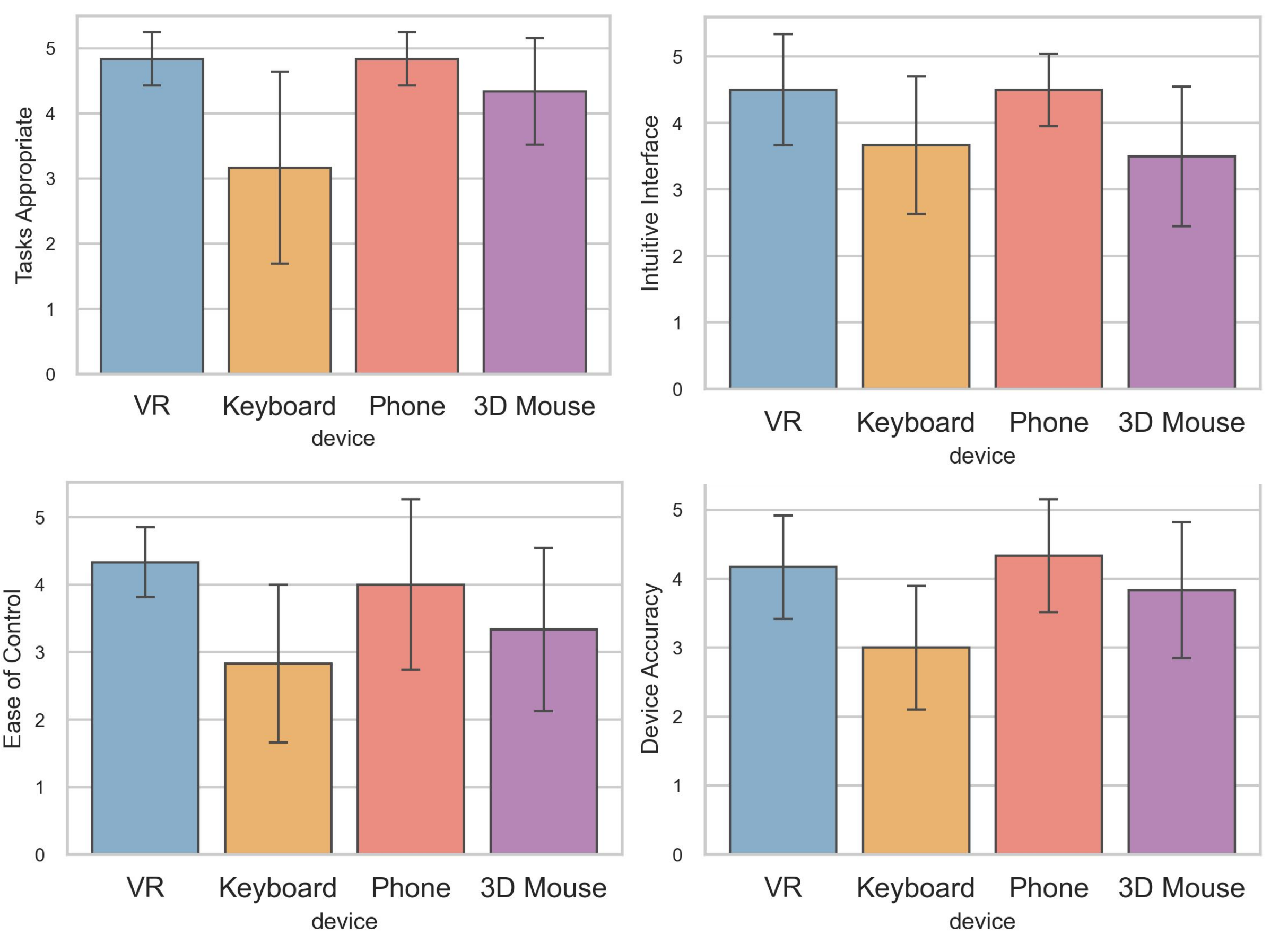}
    \caption{Additional user experience Likert scale results (Mean scores per device). Higher scores (out of 5) indicate stronger agreement with positive statements (e.g., ease of control, intuitive interface). Error bars show standard deviation.}
    \label{fig:user_experience_supp_app} 
\end{figure}

\begin{figure}[!t] 
    \centering
    \includegraphics[width=\linewidth]{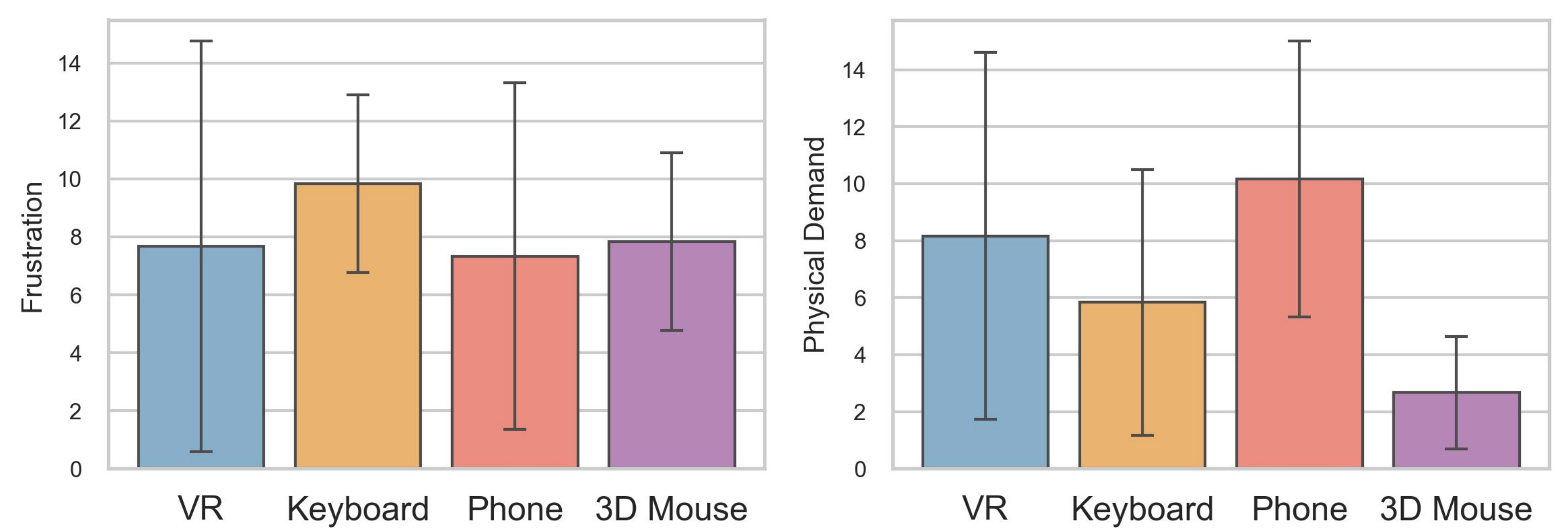} 
    \caption{User self-reported scores across different input devices via the NASA-TLX survey. Higher values indicate higher perceived workload (less favorable), except for Performance (Q4), where higher is better. Error bars show standard deviation.}
    \label{fig:nasa_tlx_app} 
\end{figure}

\begin{figure}[!t] 
    \centering
    \includegraphics[width=\linewidth]{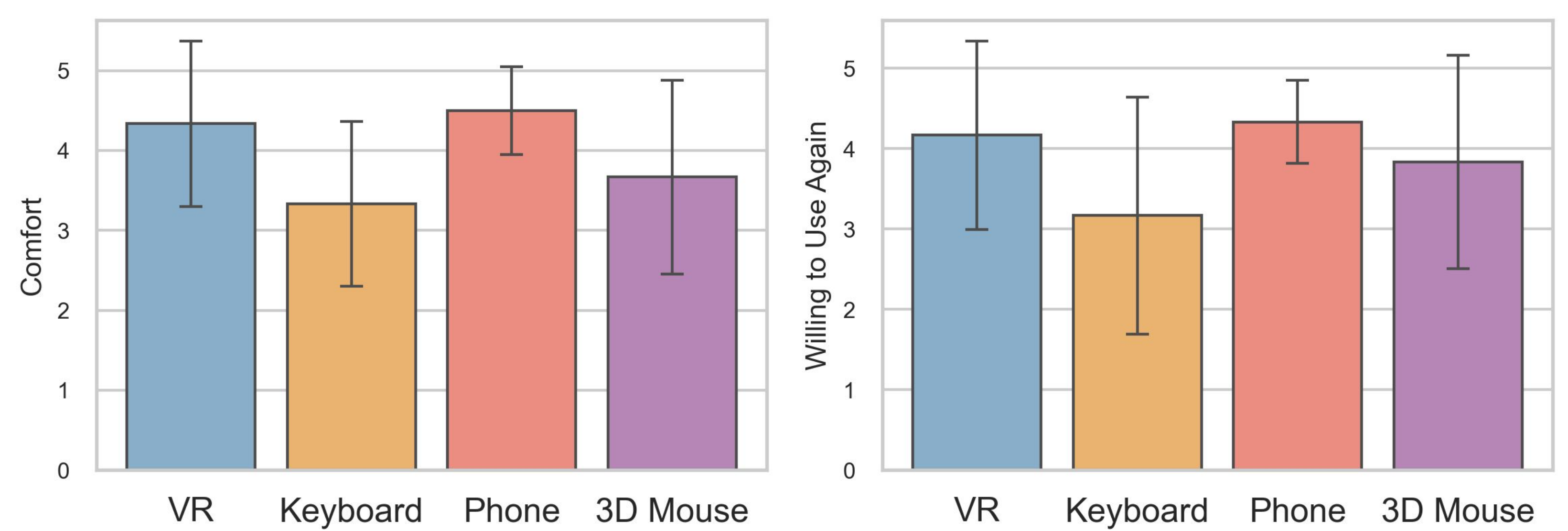} 
    \caption{User self-reported Likert scale scores across different input devices. Higher scores (out of 5) indicate a better user experience (more agreement with positive statements). Error bars show standard deviation.}
    \label{fig:user_experience_app}
\end{figure}

\end{document}